\documentclass[11pt, oneside]{article}    

\usepackage{geometry}
\usepackage{graphicx}
\graphicspath{ {./images/} }
\usepackage{color}
\usepackage{xcolor}
\usepackage{amssymb}
\usepackage{hyperref}
\usepackage{csquotes}
\usepackage{array}
\usepackage{listings}
\usepackage{pdfpages}
\usepackage{textcomp}
\usepackage{booktabs}
\usepackage{enumitem}
\usepackage{xspace}
\usepackage{subcaption}
\usepackage[multiple]{footmisc}
\setlist[description]{leftmargin=\parindent}

\usepackage[
backend=biber,
style=numeric,
sorting=none
]{biblatex}
\hypersetup{
    colorlinks=true,
    linkcolor=black,
    citecolor=black,
    urlcolor=blue,
}

\newcommand{\flatland}{\textit{Flatland}\xspace}

\title{Flatland-RL : Multi-Agent Reinforcement Learning on Trains}

\author{
  \small \textbf{\scriptsize Sharada Mohanty}\textsuperscript{*}\\
  \small \texttt{\scriptsize mohanty@aicrowd.com} \\
  \texttt{\scriptsize AIcrowd}
  
  \and
  \small \textbf{\scriptsize Erik Nygren}\thanks{These authors have equal contribution}\\
  \small \texttt{\scriptsize erik.nygren@sbb.ch} \\
  \texttt{\scriptsize SBB CFF FFS}
  \and
  \small \textbf{\scriptsize Florian Laurent}\\
  \small \texttt{\scriptsize florian@aicrowd.com} \\
    \texttt{\scriptsize AIcrowd}
  \and
  \small \textbf{\scriptsize Manuel Schneider}\\
  \small \texttt{\scriptsize manuel.schneider@hest.ethz.ch} \\
    \texttt{\scriptsize ETH ZUrich}
  \and
  \small \textbf{\scriptsize Christian Scheller}\\
  \small \texttt{\scriptsize christian.scheller@fhnw.ch} \\
\texttt{\scriptsize AIcrowd, FHNW}
  \and
  \small \textbf{\scriptsize Nilabha Bhattacharya}\\
  \small \texttt{\scriptsize nilabha2007@gmail.com} \\
  \texttt{\scriptsize AIcrowd}
  \and
  \small \textbf{\scriptsize Jeremy Watson}\\
  \small \texttt{\scriptsize jeremy@aicrowd.com} \\
  \texttt{\scriptsize AIcrowd}
  \and
  \small \textbf{\scriptsize Adrian Egli}\\
  \small \texttt{\scriptsize adrian.egli@sbb.ch} \\
    \texttt{\scriptsize SBB CFF FFS}
  \and
  \small \textbf{\scriptsize Christian Eichenberger}\\
  \small \texttt{\scriptsize christian.markus.eichenberger@sbb.ch} \\
    \texttt{\scriptsize SBB CFF FFS}
  \and
  \small \textbf{\scriptsize Christian Baumberger}\\
  \small \texttt{\scriptsize christian.baumberger@sbb.ch} \\
  \texttt{\scriptsize SBB CFF FFS}

  \and
  \small \textbf{\scriptsize Gereon Vienken}\\
  \small \texttt{\scriptsize gereon.vienken@deutschebahn.com} \\
  \texttt{\scriptsize Deutsche Bahn}
  \and
  \small \textbf{\scriptsize Irene Sturm}\\
  \small \texttt{\scriptsize irene.sturm@deutschebahn.com} \\
  \texttt{\scriptsize Deutsche Bahn}
  \and
  \small \textbf{\scriptsize Guillaume Sartoretti}\\
  \small \texttt{\scriptsize guillaume.sartoretti@nus.edu.sg} \\
    \texttt{\scriptsize Nanyang Technological University, Singapore}
    
  \and
  \small \textbf{\scriptsize Giacomo Spigler}\\
  \small \texttt{\scriptsize gspigler@uvt.nl} \\
    \texttt{\scriptsize Tilburg University}
    
}

\date{\today}
\begin{document}
\maketitle

\begin{abstract}

Efficient automated scheduling of trains remains a major challenge for modern railway systems. The underlying vehicle rescheduling problem (VRSP) has been a major focus of Operations Research (OR) since decades. Traditional approaches use complex simulators to study VRSP, where experimenting with a broad range of novel ideas is time consuming and has a huge computational overhead.
In this paper, we introduce a two-dimensional simplified grid environment called ``Flatland'' that allows for faster experimentation. Flatland does not only reduce the complexity of the full physical simulation, but also provides an easy-to-use interface to test novel approaches for the VRSP, such as Reinforcement Learning (RL) and Imitation Learning (IL). In order to probe the potential of Machine Learning (ML) research on Flatland, we (1) ran a first series of RL and IL experiments and (2) design and executed a public Benchmark at NeurIPS 2020 to engage a large community of researchers to work on this problem.
Our own experimental results, on the one hand, demonstrate that ML has potential in solving the VRSP on Flatland. On the other hand, we identify key topics that need further research. Overall, the Flatland environment has proven to be a robust and valuable framework to investigate the VRSP for railway networks. Our experiments provide a good starting point for further research and for the participants of the NeurIPS 2020 Flatland Benchmark. All of these efforts together have the potential to have a substantial impact on shaping the mobility of the future.

\end{abstract}

\subsection*{Keywords}
multi-agent reinforcement learning, operations research, vehicle re-scheduling problem, automated traffic management system, deep reinforcement learning

\section{Introduction}

The Swiss Federal Railway Company (SBB) operates the densest mixed railway traffic network in the world. On a typical day of operations, more than 10,000 train runs are executed on a network of more than 13,000 switches and 32,000 signals. Almost 1.2 million passengers and 50\% of all goods within Switzerland are transported on the railway network each day. The demand for transportation is forecasted to grow further in the next years, requiring SBB to increase the current transportation capacity of the network by approximately 30\%.

This increase in transport capacity can be achieved through different measures such as denser train schedules, large infrastructure investments, and/or investments in new rolling stock~\cite{sr40}. However, SBB currently lacks suitable technologies and tools to quantitatively assess these different measures. 

A promising solution to this dilemma is a complete railway simulation that efficiently evaluates the consequences of infrastructure changes or schedule adaptations for network stability and traffic flow. A complete railway simulation consists of a full dynamical physics simulation as well as an automated traffic management system (TMS).

The research group at SBB has developed a high performance simulator which represents and simulates the railway infrastructure and the dynamics of train traffic. The simulation is the basis for investigating the possibility of an automated traffic management system (TMS) ~\cite{egli_nygren_2018} that manages all traffic on the network by selecting train routes and deciding on the train orders at switches in order to optimize the flow of traffic through the network.

In essence, the TMS's needs to (computationally) solve the so-called vehicle scheduling problem (VSP). The VSP has been a main research topic for operations research (OR) for many decades~\cite{potvin1993parallel,foster1976integer} and was already described in detail in the early 1980s~\cite{bodin_classification_1981}. In 2007, Li, Mirchandani and Borenstein proposed the more general vehicle re-scheduling problem (VRSP)~\cite{li2007vehicle}:

\vspace{0.5cm}
\noindent \textit{The vehicle rescheduling problem (VRSP) arises when a previously assigned trip is disrupted. A traffic accident, a medical emergency, or a breakdown of a vehicle are examples of possible disruptions that demand the rescheduling of vehicle trips. The VRSP can be approached as a dynamic version of the classical vehicle scheduling problem (VSP) where assignments are generated dynamically.} 
\vspace{0.5cm}

However, solving the VRSP while taking into account all aspects of the real world that are represented in SBB's high performance physical simulation, means dealing with a NP-complete problem in an extremely complex setting: fast iteration cycles for experimenting with new ideas are impossible. In order to enable a faster pace of innovation, we propose a simpler 2D grid environment called ``Flatland'' which allows for fast experimentation with ideas and, eventually, applying the results back to the high performance physical simulation. \flatland aims to address the VRSP by providing a simplistic grid world environment. \flatland's architecture is designed to facilitate exploring novel approaches from Machine Learning Researchers, in particular from Reinforcement Learning (RL) communities where the VRSP has become of interest in recent years~\cite{larsen2019predicting,vsemrov2016reinforcement,shahrabi2017reinforcement}.

\flatland represent railway networks as 2D grid environments with restricted transitions between neighbouring cells. On the 2D grid multiple train runs have to be performed while minimizing the global delay on the network. In Reinforcement Learning terms the problem can be described as multiple agents (trains) with different objectives (schedules) who need to collaborate in order to maximize a global long-term reward.

In 2019, we ran the first version of the \flatland Benchmark \footnote{\href{https://www.aicrowd.com/challenges/flatland-challenge/}{https://www.aicrowd.com/challenges/flatland-challenge/}} a collaboration between SBB and AIcrowd, in order to test the \flatland environment and invite the AIcrowd community to tackle VRSP for the specific case of a railway system. The second \flatland Benchmark was organized as a part of NeurIPS 2020. The different tasks that were expected to be solved are explained in Section~\ref{sec:tasks}.

The key goal of \flatland is to foster innovation with respect to classical methods of Operations Research (OR), Reinforcement Learning (RL) and, importantly, combinations of OR and RL that leverage the strength of both fields, in order to provide novel solutions for the VRSP on railway networks.

In section \ref{sec:grid_world}, we summarize the concept of the \flatland environment and the implementation details. In section \ref{sec:challange} we describe design of the \flatland Benchmark and our own exploratory experiments are described in \ref{sec:methods}. We then report the results of our experiments in section \ref{sec:results} and discuss the results and the potential of the \flatland environment in the final section.

\section{The \flatland Grid World}\label{sec:grid_world}

\begin{figure}[ht]
    \centering
    \includegraphics[width=0.6\textwidth]{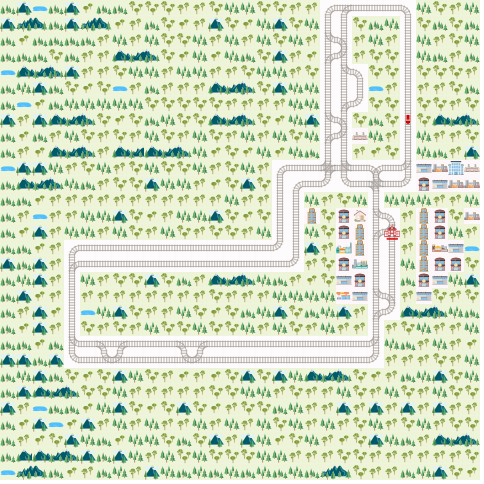}
    \caption{ \em Visualization of a simple \flatland environment with one train station and one train.}
    \label{fig:sparse_env_vis}
\end{figure}

The \flatland library is a comprehensive framework that allows to easily run VRSP experiments for railway networks. Figure \ref{fig:sparse_env_vis} shows the visualisation of a simple \flatland environment. In the following, we describe the concepts and implementation of \flatland in detail.

\subsection{Environment}
\flatland is a 2D grid environment of arbitrary size, where the most primitive unit is a \textbf{cell}. A \textit{cell} is a location in the grid environment represented by two integer coordinates $x$ and $y$, where $x \in [0, w-1]$ and $y \in [0, h-1]$, with $w$ beeing the width and $h$ the height of the grid world. Each cell has the capacity to hold a single \textbf{agent}.\footnote{In principle, the implementation of the environment allows cells to hold more than one agent, but for the sake of simplicity/realism, we assume the capacity of a single cell to be 1 in the context of the benchmark and the application to a railway system.}

An agent is an entity which is located at a particular cell, and has a direction value $d \in [0, 3]$ representing its current orientation. The direction values represent the 4 cardinal directions (North, East, South, West). An agent can move to a subset of adjacent cells. The subset of adjacent cells that an agent is allowed to transition to is defined by a \textbf{transition map}.

\flatland is a discrete time simulation, i.e. it performs all actions with constant time step. A single simulation step synchronously moves the time forward by a constant increment, thus enacting exactly one action per agent.

\subsection{Transition Maps and Emergent Patterns}\label{sec:transition_maps}

Each cell of the simulation grid has a transition map which defines the movement possibilities of the agent out of the cell. 

A cell's transition map is stored as a 4-bit bitmask that represents possible movements to the four adjacent cells given the agent's current direction. Since the movement of agents in \flatland is directional (eg. a train cannot move backwards on the track), each cell has four such 4-bit transition maps (e.g., the transition map of an east-facing agent is different than that of a south-facing agent). This setup allows to store the information about the constrained transitions using 4x4 bits of information per cell. At the same time it allows to efficiently access this data in $O(1)$ time at a minimal memory expense ($O(w \cdot h)$). For example, to store the transition map for a grid size of $1000x1000$ takes only 1.9 MB of space for the whole grid. Section~\ref{sec:transition_maps} describes in detail which movement patterns emerge from the transition maps currently used in \flatland.

\begin{figure}[ht]
    \centering
    \includegraphics[width=0.2\textwidth]{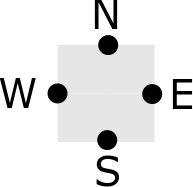}
    \caption{\em Each cell can have transitions to its $4$ neighbouring cells located North, East, South or West.}
    \label{fig:transition_maps}
\end{figure}

\begin{figure}[ht]
    \centering
    \includegraphics[width=\textwidth]{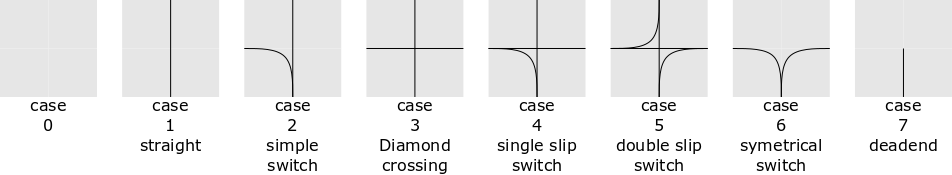}
    \caption{\em The railway configuration of \flatland consists of eight basic transition maps. Each basic transition can occur in all four rotations. A transition from one cell to another is allowed if there is a line connecting one cell to the other. Curves are a special type of case 1 transition maps where the transition includes a 90 degrees rotation. }
    \label{fig:transtition_maps_types}
\end{figure}

For this railway specific problem, $8$ basic types of transition maps (as outlined in Figure~\ref{fig:transtition_maps_types}) are defined to describe a real world railway network. All possible transitions maps are given by rotating the basic transitions maps by 0, 1, 2 or 3 directions. As in real railway networks there are never more than two transition options present in a given situation. This means that in \flatland a train needs to take at most a binary decision. Changes of agent orientation can happen for particular transitions and only if these are allowed.

\begin{description}
    \item[Case 0:] represents an empty cell, thus no agent can occupy the tile at any time. 
    \item[Case 1:] represent a passage through the cell. While on the cell the agent can make no navigation decision. The agent can only decide to either continue, i.e., passing on to the next connected tile or wait.
    \item[Case 2:] represents a simple switch. An train incoming from South in this example has to take a navigation decision (either West or North). In general, the straight transition (S $\,\to\,$ N in the example) is less costly than the bent transition. Therefore, these decisions could be rewarded differently in future implementations. Case 6 is identical to case 2 from a topological point of view but has symmetrical transition costs (both choices involve a turn).
    \item[Case 3:] can be seen as a superposition of two times Case 1. No turn can be taken when coming from any direction.
    \item[Case 4:] represents a single-slit switch. In the example, a navigation choice is possible when coming from West or South.
    \item[Case 5:] a navigation choice must be taken when coming from any direction.
    \item[Case 6:] represents a symmetrical switch with the same transition costs for both choices. At this switch, the agent must either turn left or right if coming from the South in this example. Coming from East or West, this cell acts like a simple turn.
    \item[Case 7:] represents a dead-end. Therefore, only stopping or changing direction (implemented by ``continuing forward'', see Section~\ref{sec:action}) is possible when an agent enters this cell from the South.
\end{description}

\begin{figure}[ht]
    \centering
    \includegraphics[width=\textwidth]{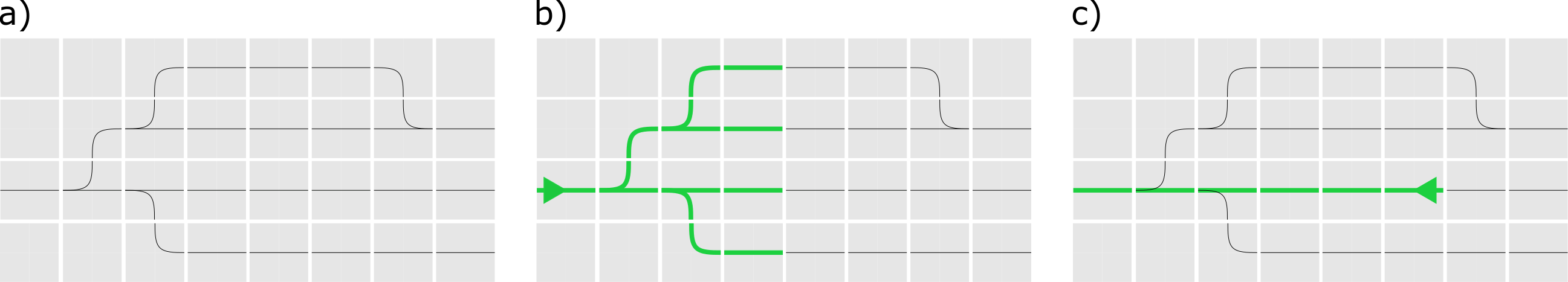}
    \caption{\em \textbf{a)} Cells can only be connected if their outgoing transitions match the incoming transitions of their neighbours. Following this rule any real world railway network can be implemented within the \flatland environment. The transition maps are direction dependent and thus the possible paths for an agent on \flatland depend not only on its position but also its current direction as shown in sub figures \textbf{b)} and \textbf{c)}.}
    \label{fig:emergent_transitions}
\end{figure}

For consistency, all outgoing connections of a cell must be joined to incoming connections of a neighboring cell. We use this consistency rule to generate random valid railway network configurations (Figure~\ref{fig:emergent_transitions}) which are used for training. We use a hold-out set of network configurations for validation and evaluation for our experiments and for the benchmark.

Agents can only travel in the direction they are currently facing. Thus, the permitted transitions for any given agent depend both on its position and on its direction (see Figure~\ref{fig:emergent_transitions}).

\subsection{Action Space}  \label{sec:action}
The action space is $discrete(5)$ for this railway specific implementation of \flatland, and consists of:
\begin{itemize}
    \item Go forward (or turn to opposite direction and continue forward if the agent is at a dead end; see Case 7 before),
    \item Select a left turn,
    \item Select a right turn,
    \item Halt on current cell, always valid, and
    \item No-op, always valid.
\end{itemize}
This definition of the action space differs from typical environments used in the RL community where agents move unconstrained and are able to change direction by rotation of the agent.

The "Go forward" action can only be executed if the transition map of the current cell allows the agent to move to the adjacent cell in its current direction. Similarly, selecting a left/right turn can only be executed when a switch is present (in the transition map of the current cell) with an allowed transition either to the left or to the right (see Figure~\ref{fig:transition_maps}; Cases 2,4,5 and 6). In cases where the turn is chosen, the direction of the agent is changed simultaneously with the move to the next cell. Finally, the "no-op action" lets the agent continue what it was doing previously: if the agent was already moving, then the no-op is equivalent to the ``go forward'' action, while a halted agent will remain stopped.

In a multi-agent setting, any action is executed if and only if it is allowed by the transition map of the current cell and if the subsequent cell will not be occupied by another agent at the next time step. If an action cannot be executed, the environment executes instead a \textit{no-op} action.

\subsection{Observations}\label{sec:observations}
This benchmark differs from many other reinforcement learning benchmarks because the objective for participants is two fold. On one hand, participants need to produce an efficient, intelligent agent which solves the VRSP tasks (Section~\ref{sec:tasks}). On the other hand, we encourage participants to design their own observation spaces in order to improve the performance of their agent. Creating this secondary focus on observation design and presenting an environment that provides maximum information for this task sets the \flatland benchmark apart from other similar competitions.

As an inspiration and baseline, we provide two different implementations of the observation space as well as access to the full information of the internal state of the environment in order to allow participants to propose and test their own representations. Furthermore, we provide wrapper functions that can be used to generate novel observation space representations by the participants.

\subsubsection{Global Observation}
The global observation is conceptually simple: every agent is provided with a global view of the full \flatland environment. This can be compared to the full, raw-pixel data used in many Atari games. The size of the observation space is $h\times w\times c$, where $h$ is the height of the environment, $w$ is the width of the environment and $c$ s the number of channels of the environment. These channels can be modified by the participants. In the initial configuration we include the following $c=5$ channels:

\begin{enumerate}
    \item \textbf{Channel 0:} one-hot representation of the self agent position and direction.
    \item \textbf{Channel 1:} other agents’ positions and directions.
    \item \textbf{Channel 2:} self and other agents’ malfunctions.
    \item \textbf{Channel 3:} self and other agents’ fractional speeds.
    \item \textbf{Channel 4:} number of other agents ready to depart from that position.    
\end{enumerate}

\subsubsection{Tree Observation}
\begin{figure}[ht]
    \centering
    \includegraphics[width=\textwidth]{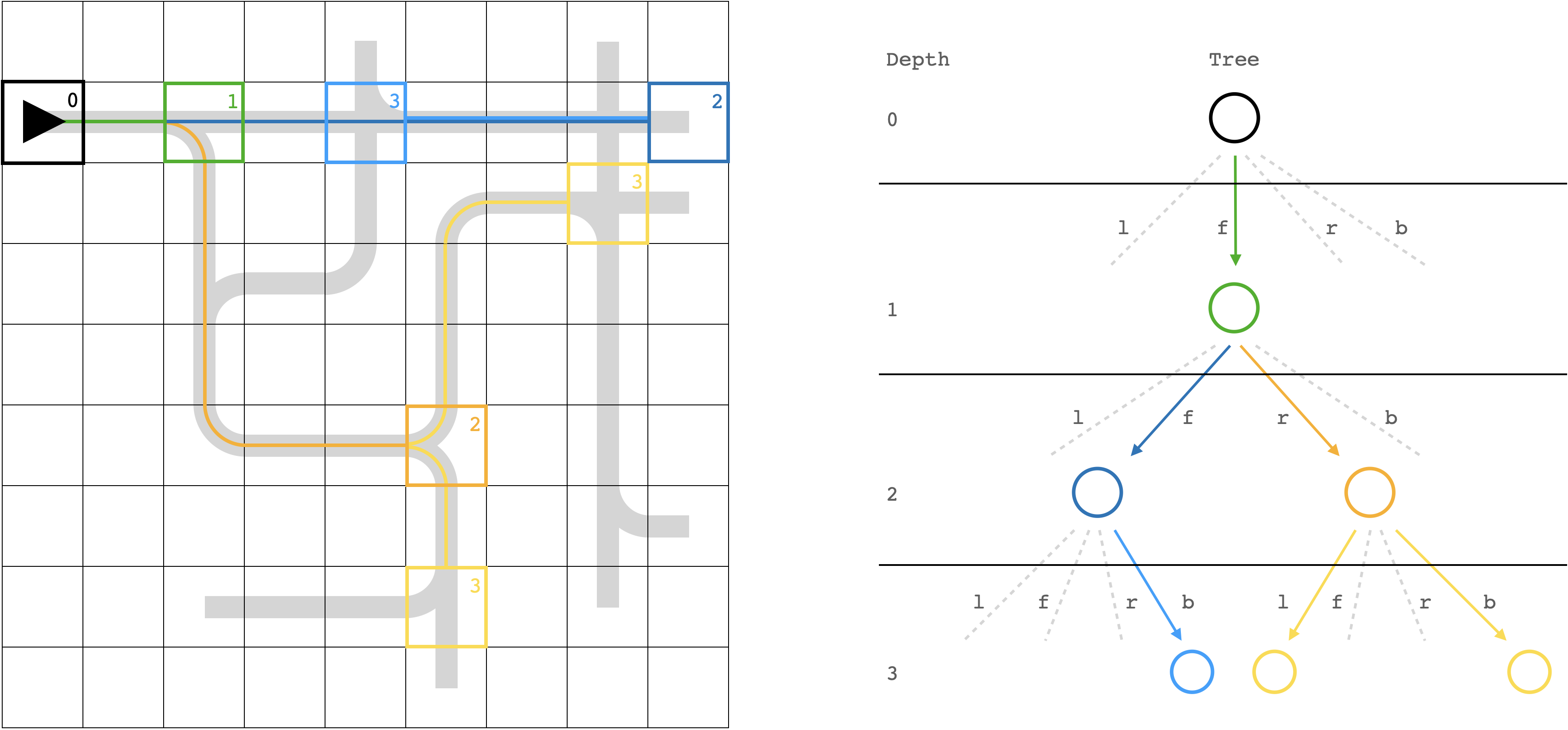}
    \caption{ \em Building a tree observation of depth 3. The branches are built in the order left, forward, right, and backward of the agent's current orientation. Only valid transitions result in a branch whereas backward is only allowed in a dead end. The right side of the figure shows the tree corresponding to the railway network on the left, where the triangle depicts the agent and its orientation. In the tree, a circle indicates a node populated with the corresponding information.}
    \label{fig:tree_search}
\end{figure}

The tree observation is built by exploiting the graph structure of the railway network. The observation is generated by spanning a four-branched tree from the current position of the agent. Each branch follows the allowed transitions (backward branch only allowed at dead-ends) until a cell with multiple transitions is reached, e.g. a switch. The information gathered along the branch until this decision point is stored as a node in the tree. In more detail, the tree observation is constructed as follows (see also Figure~\ref{fig:tree_search}):

From the agent's location,
\begin{enumerate}
    \item probe all 4 directions starting with left and start a branch for every allowed transition.
    \item Walk along the track of the branch until a dead-end, a switch or the target destination is reached,
    \item Create a node and fill in the node information as stated below.
    \item If max depth of tree is not reached and there are possible transitions, start new branches and repeat the steps 1 to 4.
    \item Fill up all non existing branches with -infinity such that tree size is invariant to the number of possible transitions at branching points.
\end{enumerate}

Note that we always start with the left branch according to the agent orientation. Thus, the tree observation is independent of the orientation of cells, and only considers the transitions relative to the agent’s orientation.

Each node is filled with information gathered along the path to the node. Currently each node contains 9 features:

\begin{enumerate}
    \item if own target lies on the explored branch, the current distance from the agent in number of cells is stored.
    \item if another agent’s target is detected, the distance in number of cells from the current agent position is stored.
    \item if another agent is detected, the distance in number of cells from the current agent position is stored.
    \item conflict detected on this branch
    \item if an unusable switch (for the agent) is detected we store the distance. An unusable switch is a switch where the agent does not have any choice of path, but other agents coming from different directions might.
    \item This feature stores the distance (in number of cells) to the next node (e.g. switch or target or dead-end)
    \item minimum remaining travel distance from this node to the agent’s target given the direction of the agent if this path is chosen
    \item agent in the same direction found on path to node
    \begin{itemize}
        \item n = number of agents present in the same direction (possible future use: number of other agents in the same direction in this branch)
        \item 0 = no agent present in the same direction
    \end{itemize}
    \item agent in the opposite direction on path to node
    \begin{itemize}
        \item n = number of agents present in the opposite direction to the observing agent
        \item 0 = no agent present in other direction to the observing agent
    \end{itemize}
\end{enumerate}

Figure~\ref{fig:tree_search} illustrates the tree observation on a simple network where the switches are highlighted and numbered in order to identify them with their corresponding nodes in the tree.

\subsection{Reward}\label{sec:rewards}

Each agent receives a combined reward consisting of a local and a global reward signal. For every time step $t$, an agent $i$ receives a local reward $r_l^i = -1$ if it is moving or stopped along the way and $r_l^i = 0$ if it has reached its target location. In addition, a configurable penalty $r_p^i$ is received if the move is illegal, i.e. the action cannot be executed. This penalty is set to $0$ in the NeurIPS 2020 benchmark. If all agents have reached their targets at time step $t$, a global reward $r_{g} = 1$ is awarded to every agent. If not all the agents is not at its destination, all agents get a global reward $r_g^t = 0$. Overall, every agent $i$ receives the reward
\begin{equation}
    r_{i}(t) = \alpha \, r_l^i(t) + \beta \, r_g(t) + r_p^i(t) \hspace{1cm}\in{[-\alpha-2,\beta]},
\end{equation}
where $\alpha$ and $\beta$ are parameters for tuning collaborative behavior. Therefore, the reward creates an objective of finishing the episode as quickly as possible in a collaborative way. This structure is similar to that of existing multi-agent path finding RL casts~\cite{sartoretti2019primal}.

At the end of each episode, the return of each agent $i$ is given by the sum over the time step rewards:
\begin{equation}
    g_i = \sum _{t = 1}^T r_i(t),
\end{equation}
where $T$ is the time step when the episode terminated.

\section{The \flatland Benchmark}\label{sec:challange}
The \flatland library is only a means to solve the VRSP at hand. A public benchmark offers a great opportunity to collectively tackle this difficult problem. Some specifics of NeurIPS 2020 \flatland Benchmark are described below, please visit the dedicated website\footnote{\href{https://www.aicrowd.com/challenges/neurips-2020-flatland-challenge/}{https://www.aicrowd.com/challenges/neurips-2020-flatland-challenge/}} for more information.

\subsection{Related Benchmarks}
There have been other multi-agent reinforcement learning benchmarks where the objective was to learn either collaboration or competition (or both) among agents, e.g., Pommerman~\cite{pommerman}, Marl\"{o}~\cite{marlo}, SC2LE~\cite{starcraft}. The key aspects in which the \flatland benchmark differs from the previous benchmarks are:
\begin{itemize}
    \item In \flatland, the tasks have to be solved by a large group of agents ($n>100$), which means that collaboration becomes a much more complex problem. These coordination tasks cannot be solved by relying just on local information (e.g. limited view of agents). At the same time, scalability constraints limit the use of centralized approaches, thus introducing the need for subtle new observation spaces and training methods.

    \item The allowed transitions in \flatland are much more constrained than free movement on a grid. Participants are therefore forced to explore novel approaches for representing the observation spaces in contrast to pixel or grid view observations used in other environments. Furthermore, the distribution and importance of actions in \flatland's action-space is highly imbalanced. In many situations only a single action is possible (e.g., go-forward). Much less frequently, there is more than one option and the decisions are highly critical to the task's success (e.g., decisions to turn at junctions). This imbalance results in more difficult exploration at training time, and will likely require the need for more complex training methods to allow agents to efficiently explore the state-action space, as well as to identify and reason about key decisions along their way.

    \item \flatland directly addresses a real world problem. Solutions submitted to the benchmark have a practical impact on issues such as real-time railway re-scheduling. The VRSP belongs to a wider class of collaborative, scalable multi-agent cooperation tasks, which includes the job shop scheduling problem~\cite{jo1999survey}, the aircraft landing problem~\cite{zi2018review}, and many more problems connected to logistics and transportation. Therefore, any promising solution found in context of this benchmark, will be applicable in a broad range of industries, where improved autonomous collaboration could have a large impact on society.

    \item Along the same lines, the existing \flatland (open-source) code base provides an extensible framework to define novel tasks beyond railway scheduling problems. An adaptation of the \flatland could lead to easy and accessible tools for studying other classes of related optimization problems, e.g. optimization of ground traffic at airports in challenges and applications.
\end{itemize}

Typically, (Multi Agent) RL benchmarks are tailored to match specific research questions and therefore present a fixed design of observation space and reward functions. In contrast, \flatland exposes the full internal state of the environment, presenting participants with larger freedom. This enables them to apply their creativity also to the design of observations, an aspect that has been found to be essential for the success of RL solutions and that is worth to be incorporated into the core library.

\flatland also builds upon features of many successful Multi-Agent Reinforcement Learning Environments. Participants can interact with the environment using a familiar gym-flavoured API inspired by numerous other Multi Agent RL environments.

\subsection{Task and rounds}\label{sec:tasks}
The NeurIPS 2020 \flatland benchmark introduces the VRSP to the participants in two main rounds. In both rounds, the agents have to navigate from a starting cell to a target cell at a constant speed. In addition, agents can be inflicted by malfunction on their run. The occurrence of malfunctions is defined in the environment by a malfunction probability and a malfunction duration interval.

Each round introduces a separate set of test environments and a different set of constraints.

The goal of the first round was to maximise the accumulated reward for solving a set of 14 tests, each including several evaluation environments. Prerequisite for a valid submission was that all evaluation environments had to be solved within 8 hours, otherwise the submission failed and no score was awarded.

In the second round, the goal was not only to maximize the score for the test environments, but also to minimize computation time. This goal directly relates to the business requirements of real world railway operations that require short response times in case of malfunction/disruptions. Submissions to round two had to solve a open-ended sequence of test environments, starting with small grid sizes and small number of trains with progressively increasing grid sizes and number of trains. The goal was to solve as many of these environments as possible in 8 hours. The evaluation was stopped pre-maturely if less than 25\% of the agents reached their target, averaged over each test of 10 episodes.

\subsection{Custom Observations}

Our own experiments showed that the design of the observation space is a crucial element for all optimal solutions. Therefore, participants were given the flexibility to design their own observations, either by modifying and combining \flatland's ``stock'' observations or using the full \flatland state information and predefined wrapper functions.

\subsection{Metrics}\label{sec:metrics}
For each task, the agents submitted by the participants were evaluated on a fixed set of environment configurations ordered from low to high complexity. The solutions were evaluated using the normalized reward per agent across all environments, which is calculated as follows :
\begin{equation}
   S = \sum_{j = 0}^{N} \sum_{i = 0}^{M_j} g_{i}^j. \label{eq:score}
\end{equation}
where the score ($S$) is the accumulated total reward (Section~\ref{sec:rewards}) for $N$ completed environment configurations with $M_i$ agents for the respective configuration. The participants' solutions were ranked by the score $S$ where a higher score is better.

\subsection{Tutorial and documentation}

The \flatland project website\footnote{\url{http://flatland.aicrowd.com}} features a documentation for the \flatland environment with a step by step introduction to the core components and API references, information on the ongoing and past benchmarks, information on research approaches, and Frequently Asked Questions (FAQ). The code for the environment, provided under an open-source license, is available for the participants in the public \flatland repository.\footnote{\url{https://gitlab.aicrowd.com/flatland/flatland}}

Participants can make their first submission by following the instructions in the benchmark starter kit.\footnote{\url{https://gitlab.aicrowd.com/flatland/neurips2020-flatland-starter-kit}}. Beyond the starter-kit, we also provide baseline implementations which can act as a good starting point for participants.

\section{Experiments on \flatland}\label{sec:methods}

\flatland provides the basis to use RL to tackle VRSP problems. As a first step towards demonstrating the potential of RL for this problem, we conducted a series of experiments on \flatland. We evaluated a number of well-known RL algorithms in standard and customized versions and applied additional alterations to problem formulation and observation. Our results provide first insights into the performance and the key challenges of RL for VRSP.

The experiments are built upon the RLlib framework~\cite{liang2018rllib}, which provides a diverse set of state-of-the-art reinforcement learning algorithms such as PPO~\cite{zoph2018learning}, IMPALA~\cite{pmlr-v80-espeholt18a} and Ape-X~\cite{horgan2018distributed}. We include results from:
\begin{itemize}
    \item{Random agents, shortest-path agents and solutions based on OR alternatives as baselines}
    \item{Standard and customized RL algorithms:Standard PPO and Ape-X, PPO with centralized critic (CCPPO) with custom observations and/or variants of action handling }
    \item{Imitation learning approaches }
\end{itemize}

With Ape-X using a DQfD loss, we evaluate how expert demonstrations from OR solutions can be used for imitation learning. Compared to other imitation learning algorithms such as DAGGER~\cite{ross2010reduction} and Deeply AggreVaTe~\cite{sun2017deeply}, a key advantage of DQfD~\cite{hester2018deep} is the ability to improve beyond the demonstrations which is essential to master large \flatland instances in which OR approaches do not scale.

\subsection{Training, Evaluation and Test Protocols}\label{sec:protocols}

For all our experiments, we use the sparse rail generator\footnote{\url{https://flatland.aicrowd.com/getting-started/env/level_generation.html\#sparse-rail-generator}} from the \flatland library to generate the environments. All environments had a size of $25 \times 25$ cells and 5 agents. The detailed configurations for the generator and the experiments can all be found in the baseline repository.\footnote{\url{https://gitlab.aicrowd.com/flatland/neurips2020-flatland-baselines/blob/flatland-paper-baselines/envs/flatland/generator_configs/small_v0.yaml}}\footnote{ \url{https://gitlab.aicrowd.com/flatland/neurips2020-flatland-baselines/tree/flatland-paper-baselines/baselines}}

Training for pure RL algorithms like Ape-X and PPO were done for 15 million steps. For Imitation Learning (IL) algorithms, which were run with saved expert experiences, more training steps were run. MARWIL was run for 5 billion steps whereas Ape-X FIXED IL(25\%) was run for 1 billion steps. Ape-X FIXED IL(100\%) was run for 100 million steps. The number of steps for these IL experiments are different as we saw significant improvement after 15 million steps. Moreover, the expert experiences were saved separately for each agent and each agent's experience is considered as a step. Hence, for an environment with 5 agents, one episode step in a typical MARL setting would be equivalent to 5 steps in the MARWIL setting. For pure IL, we only report evaluation results. Expert samples for IL were taken from the OR solution of 2019 Flatland Benchmark, by Roman Chernenko\footnote{\url{https://bitbucket.org/roman_chernenko/flatland-final}} which showed a 100\% completion rate on the tested environments.

Training was done with 3 different environment seeds. For evaluation, we ran 50 evaluation episodes using held out random seeds for every 50 training iterations. We took the checkpoint corresponding to the highest normalised reward on the evaluation set starting at 1 million steps to avoid good results due to lucky initialisation and easier environments in the beginning.

We then ran the model with this checkpoint on a separate set of 50 episodes and report this as the test set. The test set is run against 3 different training checkpoints for each algorithm. For Ape-X based runs, we sampled a random action from the action space with probability 0.02 to allow for some exploration in the experiment. The test result is shown as the mean and standard deviation of these 3 runs corresponding to the 3 different training seeds in section~\ref{sec:results}.

\subsection{Baselines}
To establish some conventional baselines for our experiments, we used:

\begin{itemize}
    \item{Stochastic random agents (Random)}
    \item{Agents that always choose the ``go forward'' action (Constant Go Forward Action)}
    \item{Agents that use the shortest Path (Shortest Path)}
\end{itemize}

\subsection{Standard and customized RL algorithms}

Our RL experiments were performed using
\begin{itemize}
    \item{Standard Ape-X and PPO with the local tree search observation.}
    \item{PPO with centralized critic (CCPPO) in a base variant and using a transformer (details see below)}
    \item{PPO and/or Ape-X with frame skipping, action masking or a global density observation (details see below)}
\end{itemize}


\subsubsection{Centralized Critic}
The \flatland environment reflects one challenge of real-world railway planning very strongly: trains need to cooperate in sharing sections of railway infrastructure. Therefore \flatland presents a challenge to learning algorithms with respect to the agent’s behavior when faced with the presence of other agents. Centralized critic methods address the problem of cooperation in multi-agent training situations with partial observability. They can be regarded as a middle ground between individual train agents that regard other trains as part of the (thus non-stationary) environment, and a single agent for all trains whose action space is the joint action space of all train agents, a solution which has serious drawbacks when it comes to scalability.

This centralized critic approach was first implemented and evaluated in a project of Deutsche Bahn in cooperation with InstaDeep. In the context of that project, CCPPO was validated and directly compared to standard PPO on a previous version of \flatland. That original implementation that used a custom version of a sparse reward and a custom observation similar to \flatland's tree observation outperformed standard PPO which motivated the current experiments with CCPPO.

The version implemented here follows the actor-attention critic approach~\cite{iqbal2019actor} and uses a centrally computed critic that learns based on the observation of all agents, but that trains decentralized policies. In addition to a basic version where the input of the centralized critic is just a combination of all agents' observations, we also provide a Transformer as proposed by Parisotto et al.
~\cite{parisotto2019stabilizing} that serves as an attention mechanism that learns to reduce the concatenated observation to the relevant information.

The base architecture implemented here is a a fully connected network trained using PPO. At execution time the agents will step through the environment in the usual way. During training, however, a different network is used that provides the evaluation of the current state on which to train the agent’s network. This ``central critic'' is provided with a more holistic view of the simulation state in order to better approximate an adequate reward for the agent’s actions in the current situation. Both network consist of three hidden layers of size 512.

In the basic version, the input to the central critic is a concatenation of all trains' observations in a consistent order. This leads to an observation that contains no information about the position of the trains in the network, nor about their relative positions. However, it provides a good starting point for experimenting with the input for the critic.

In the version using the transformer, an additional processing layer is added between the concatenated observations of the agents and the centralized critic, in order to provide the critic with an embedding of the observations. The transformer is built with self-attention layers that calculate the embedding with trained linear projections. The dimensions of these outputs can be decided: their number is the product of the number of parallel heads and of the output dimension. The transformer in the current implementation represents an attention mechanism that learns the best representation of the concatenated observations.

For the experiments described in this paper, we the used the local tree search observation and standard reward.

\subsubsection{Frame Skipping and Action Masking}
These modifications were introduced in order to ease the learning for RL agents:

\subparagraph{Skipping “no-choice” cells: }
Between intersections, agents can only move forward or stop. Since trains stop automatically when blocked by another train until the path is free, manual stopping only makes sense right before intersection, where agents might need to let other trains pass first in order to prevent deadlocks. To capitalize on this, we skip all cells, where the agent is not on or next to an intersection cell. This should help agents in the sense that they no longer need to learn to keep going forward between intersections. Also skipping steps shortens the perceived episode length.

\subparagraph{Action masking}
The \flatland action space consists of 5 discrete actions (no-op, forward, stop, left, right). However, at any given time step, only a subset of all actions is available. For example, at a switch with a left turn and a straight continuation, the agent cannot turn right. Furthermore, the noop-action is never required and can be removed completely. We test masking out all invalid and noop actions so that the agents can focus on relevant actions only.

We evaluated skipping “no-choice” cells for both DQN Ape-X and PPO agents. Action masking was only tested with PPO agents.
All experiments were performed with the stock tree observations. For policy and value functions, we employed a small two layer neural network (each with 256 hidden units and ReLU activation).

\subsubsection{Global Density Map}
Since in preliminary experiments the standard global observation didn't perform well, we experimented with alternative global observations.

The global density map observation is based on the idea that every cell of the environment has a value representing an “occupation density”. For simplicity, we assume that an agent follows the shortest path to its target and doesn't consider alternative paths. For every cell of an agent's path the distance in steps of the current position of the agent to this cell is calculated. These values are aggregated for all agents into a estimated occupation density measure. This measure encodes how many agents will likely occupy that cell in the future. The contribution of an agent to the density value is weighted by the agent's distance to the cell.

For example, if all $M$ agents' paths pass the same cell at the same time step $t_i$, the density is high. If the agents pass the same cell at different time steps $\{t_{j}^m | 0 < m \leq M, j > i \}$, the density is lower.

The occupation density value for a cell and agent is defined as:

$$d(t) = \exp(-\frac{t}{\sqrt{t_\textrm{max}}})$$

where $t$ is the number of time steps for the agent to reach the cell and $t_\textrm{max}$ the maximum number of time steps included in the density map. The density map present the agents the opportunity to learn from the (projected) cell occupancy distribution.

The observation for each agent consists of two arrays representing the cells of the environment. The first array contains the density values for the agent itself including the agent's own position, and the second array contains the mean of the other agents’ values for each cell.

\subsection{Imitation Learning}
The \flatland environment presents two major challenges to learning algorithms. On the one hand, agents not only have to learn to efficiently navigate to their target, but also to deal with path conflicts with other agents. On the other hand, rewards are sparse and delayed which renders their assignment to the relevant actions complex. OR solutions which follow rule-based and planning approaches have been shown to perform well in this environment. We used imitation learning to take advantage of the OR capabilities while potentially generalizing to larger instances than those considered in the demonstrations. 

We considered two different, broad Imitation Learning approaches:

\subsubsection{Pure Imitation Learning}
This approach involves training in a purely offline process from stored expert experiences or training online by learning to mimic actions generated by an expert.

For the offline process using stored experiences this is implemented using the RLLib supported MARWIL~\cite{NIPS2018_7866} algorithm which is an on-policy algorithm as well as the off-policy Ape-X algorithm.

We also developed a simple online training process to train a neural network model that mimics expert actions that are generated as the actual environment is run to overcome limited samples stored offline.

\subsubsection{Mixed Learning}
This step involves training in both an offline process via stored experiences and actual environment simulation experiences. This is implemented using the off-policy Ape-X algorithm with the stored experiences sampled from the replay buffer based on a user configurable ratio of simulation to expert data. To combine the losses from both the expert and the RL algorithm, we used a loss function similar to DQfD~\cite{hester2018deep}.

We also used the simple online training process in the pure imitation learning step along with the on-policy PPO algorithm~\cite{schulman2017proximal}. This was done by running entire episodes with either a Pure IL approach or a PPO Approach. The algorithm to be run was selected based on the probability (this is user configurable though for our experiments we used 50\%) of a binomial distribution. This has been adapted from the approach described in the Multi-agent path finding (MAPF) framework PRIMAL~\cite{sartoretti2019primal}.

The aim of the 2-step approach was to see the limit of what can be learned using only expert data and then improve on that using a hybrid approach to bootstrap the Reinforcement Learning process. For each of the two approaches, both, on- and off-policy RL algorithms were run. The results can be directly compared to the Ape-X and PPO runs that were also part of our experiments. 

\section{Results}\label{sec:results}

\begin{figure}[t]
    \begin{subfigure}[h]{0.49\linewidth}
        \includegraphics[width=\linewidth]{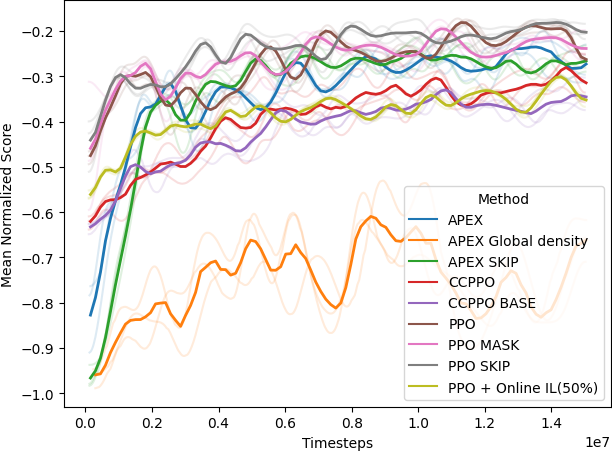}
    \end{subfigure}
    \hfill
    \begin{subfigure}[h]{0.49\linewidth}
        \includegraphics[width=\linewidth]{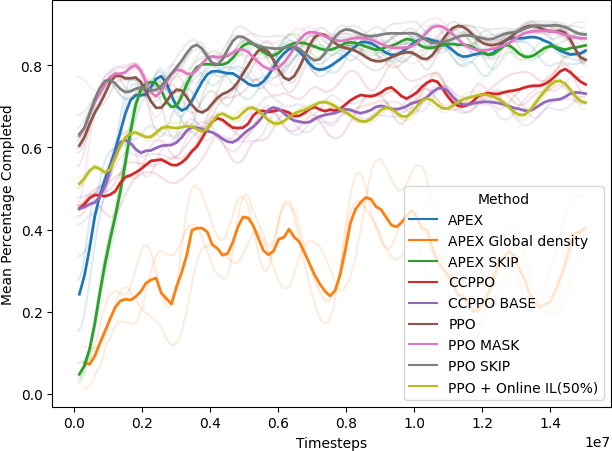}
    \end{subfigure}
    \caption{Mean normalized score (left) and completion rate (right) of the RL experiments. The figures show the mean across the respective training runs for each experiment, individual runs are displayed as fine lines. We applied a Gaussian filter for smoothing.}
    \label{fig:training_rl}
\end{figure}

\begin{figure}[t]
    \begin{subfigure}[h]{0.49\linewidth}
        \includegraphics[width=\linewidth]{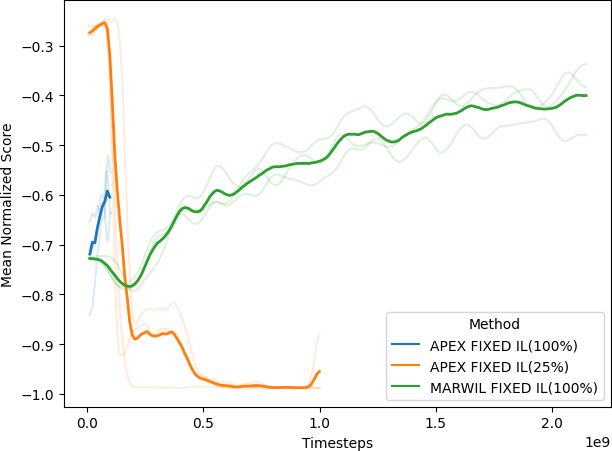}
    \end{subfigure}
    \hfill
    \begin{subfigure}[h]{0.49\linewidth}
        \includegraphics[width=\linewidth]{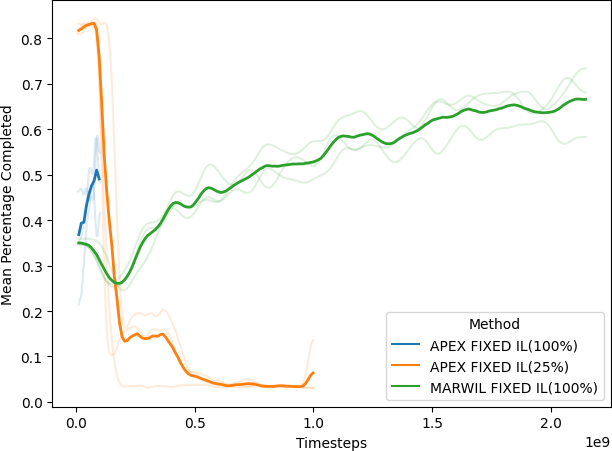}
    \end{subfigure}%
    \caption{Mean normalized score (Left) and completion rate (Right) of the IL experiments. The figures show the mean across the respective training runs for each experiment, individual runs are displayed as fine lines. We applied a Gaussian filter for smoothing.}
    \label{fig:training_il}
\end{figure}

Figure~\ref{fig:training_rl} and Figure~\ref{fig:training_il} show the training progress (per approach) averaged over three runs with different seeds.\footnote{These figures and more are available on Weights \& Biases: \url{https://wandb.ai/aicrowd/flatland-paper/reports/Flatland-Paper-Results--VmlldzoxOTc5Mjk}} The training and test results are listed in Table~\ref{tab:RLLibBaselineResults}. Refer to Section~\ref{sec:protocols} for details of the training, evaluation and test protocols).

In summary, all RL and IL experiments outperform the ``Random'' agents and ``Always Forward'' agents. Few experiments (Ape-X FIXED IL (25\%), PPO SKIP) show small improvements over the Standard Ape-X and PPO models. The Shortest-Path-Agents perform better than the low-performing RL approaches quite well (local grid observation and global density map observation) demonstrating the strength of even a basic planning approach. In the following section, more details for the individual experiments are provided. 

\subsection{Reinforcement Learning}

For the centralized critic, the results show that the base network version outperformed the transformer variation in completion rate with 76.27\% compared to 71.87\%. The base network also shows a higher standard deviation. However, in our experiments CCPPO BASE was outperformed by both the Ape-X implementation and several PPO versions. Even though, one of the best submission to the NeurIPS 2020 \flatland benchmark is built on the CCPPO implementation of our experiments.

The experiments which skip ``no-choice'' cells and mask unavailable actions show comparable performance and sample-efficiency to their standard Ape-X and PPO counterparts. Notably, The skipping of ``no-choice'' cells with Ape-X was successfully used in a submission to the NeurIPS 2020 \flatland benchmark using different hyperparameters.

The global density observation performs slightly better than the stock global observation, illustrating that there is potential in carefully designing global observations. However, the test performance is significantly lower than the training performance which warrants further investigation.

\subsection{Imitation learning}


The results show that a pure Imitation Learning can help push the mean completion to 80\% on the test results \ref{tab:RLLibBaselineResults}. This is achieved via a simple neural network that mimics the expert actions. The MARWIL algorithm performs similarly with a completion of 72.4\%. The Ape-X based pure off-policy based IL algorithm performs poorly with only 10\% completion. However supplementing this approach with 75\% samples from the actual environment and using a loss function similar to DQfD, leads to 86\% completion which is a significant improvement with respect to the best performing pure IL algorithm.  

When the simple neural network is mixed with a PPO, we see a drop in performance to a completion rate of 71.47\%. Since this mixed approach was trained only for 15 million steps and with a constant probability of IL/RL ratio, this performance could be improved by using a decayed IL/RL ratio and training for higher number of steps.

Combining both, the expert training and environment training with the fast Ape-X or the PPO based mixed IL/RL algorithm, leads to a mean completion rate comparable and higher than corresponding pure RL runs. A notable observation is that Imitation Learning algorithms have a higher minimum completion rate.

\begin{table}[t]
    \centering
    \small
    \begin{tabular}{lcccc}
        \toprule
\textbf{METHOD} & \multicolumn{2}{c}{\textbf{Train}} & \multicolumn{2}{c}{\textbf{Test}} \\ \midrule
 & \textbf{\% Complete} & \textbf{Reward} & \textbf{\% Complete} & \textbf{Reward} \\
 \midrule
Ape-X FIXED IL(25\%) & 90.45±0.4 & -0.18±0 & 86±1.44 & -0.22±0.01 \\
Ape-X FIXED IL(100\%) &  &  & 22.93±11.83 & -0.84±0.08 \\
Ape-X & 90.38±1.64 & -0.2±0.02 & 80.93±5.45 & -0.32±0.04 \\
Ape-X SKIP & 89.51±1.09 & -0.21±0.01 & 79.73±0.92 & -0.33±0.01 \\
CCPPO & 87.72±2.37 & -0.2±0.02 & 71.87±3.7 & -0.35±0.03  \\
CCPPO BASE & 83.21±1.47 & -0.25±0.01 & 76.27±6.96 & -0.31±0.06 \\
MARWIL FIXED IL(100\%) &  &  & 72.4±3.27 & -0.35±0.02 \\
PPO + Online IL(50\%) & 83.46±1.09 & -0.23±0.01 & 71.47±4.01 & -0.35±0.04 \\
PPO & 94.78±0.29 & -0.13±0.01 & 81.33±5.86 & -0.26±0.05 \\
PPO MASKING & 93.4±0.27 & -0.15±0 & 80.53±9.59 & -0.28±0.09 \\
PPO SKIP & 93.48±0.66 & -0.15±0.01 & 82.67±5.79 & -0.26±0.05 \\
Ape-X Global density & 57.87±1.85 & -0.51±0.01 & 34.4±9.23 & -0.71±0.07 \\
Online IL(100\%) &  &  & 80±3.27 & -0.27±0.03 \\
\midrule
Random &  &  & 20.4 & -0.85 \\ 
Constant Forward Action &  &  & 22.4 & -0.8 \\
Shortest Path &  &  & 67.2 & -0.38 \\
\bottomrule
\end{tabular}
    \caption{\em Evaluation and test results of the experiments.}
    \label{tab:RLLibBaselineResults}
\end{table}{}

\section{Discussion}

Our experiments demonstrate that RL agents are able to learn strategies for solving most of the test environments. Although there is a gap between the train and test mean completion rate, most completion rates range between 71 and 86\% for the test environments, demonstrating that RL models are able to generalize from the training samples to unseen problem instances. Even though the RL results presented here demonstrate the feasibility of the approach, RL solutions are clearly outperformed by OR methods that dominate the public benchmark's overall leader board. 

We note that standard Ape-X and PPO with tree observations perform as well as, or better, than other models using the same observation. There is no visible difference in sample efficiency between these experiments. Therefore, the chosen RL approaches are capable to identify the relevant information from the observation alone. Unexpectedly, PPO with the centralized critic (CCPPO) performed slightly worse than the standard PPO counterpart. This means that the assumed benefit of the centralized critic with it's global view in the training could not be proven, but on the contrary seems to affect generalization. This is supported by the finding that the basic CCPPO has a smaller drop between training and testing performances than the transformer CCPPO. This could mean that the attention mechanism makes the CCPPO's policy more specific to the states it encountered. In summary, the principle of the centralized critic that has been introduced in a basic version, needs further experimentation in combination with other models beyond PPO.

Similarly, providing the agents with the global view of the global density map leads to significantly lower completion rates. This approach outperformed the baseline agents with random and constant forward action, but performed worse than a simple planning approach given by the shortest path baseline. It is also noteworthy that the training performance oscillated much more than the approaches using the tree observation. A tentative explanation of the lower performance and the lack of stability during training could relate to the fact that the global density experiments contain a representation of the track layout of the respective environment. Therefore, the agents are likely to "memorize" the layout of the environments they were trained on rather than being able to generalize to new environment. A possible improvement could be to build the global density observation in a way that it abstracts from the track layout.

The approach using a combination of IL and RL (fixed 25\% IL) shows the highest mean completion rate in the test set with only a minor drop from the training performance. This could be due to the fact that it could learn from many successful demonstrations (the IL part) and that all actions in a demonstration are optimal with regard to the original OR strategy used to generate the demonstrations.

An obvious shortcoming of the RL approach seems to be the lack of coordination between trains. Although the tree observation provides some information about possible collisions with other agents, the agents do not seem to be able to coordinate a response to that information, resulting in frequent dead-lock situations. A way to improve the coordination between agents could be to add communication, for instance enabling a mechanism that allows agents to communicate their intended next action.

\subsection{Limitations}

A common question concerns the transferability of solutions from \flatland's simplified grid environment to real railway networks and real world railway planning tasks.

In \flatland, the movement physics of trains is simplified as much as possible: Trains either travel at full speed or do not move. Even though this may seem highly unrealistic, such simplifications are common in today's scheduling and re-scheduling methods. Today, all planning is done using simplified calculations of minimal travel-time with  additional buffer times. Even though these models are still more complex than the one used in Flatland, the general idea of planning with simplified models and additional buffers is the same.

\flatland's core principle of mutual exclusive occupancy of grid cells directly reflects the structure of signalling and safety systems: in today's railway systems the railway network is divided into discrete blocks. Each block can only be occupied by a single train. First and foremost this is a safety principle that is ensured by axle counters and signals. However, this structure also simplifies planning. In the process of digitization and automation the block signaling techniques will be replaced by the ``continuous'' moving block techniques at some point. However, the discrete problem reflects the vast majority of today's railway networks and also is relevant for discrete modelling of the ``continuous'' moving block scenario.

The patterns of possible transitions between grid cells were modelled directly after the real railway infrastructure. All transition maps of the \flatland environment reflect real infrastructure elements of the Swiss Railway Network, such as single slip switches, double slip switches, and crossings.

Exploiting the capacity of the railway infrastructure is one of the most challenging aspects of railway planning in all railway networks due to the ever increasing demand for service and to the high effort of building new infrastructure. \flatland directly reflects these aspects by presenting problems with a large ratio of single-track sections that are one source of complexity for the re-scheduling in daily railway operations. Furthermore, \flatland's problems with high numbers of trains that fill the railway network directly address the problem of handling traffic loads close to a network's maximum capacity.

\subsection{Conclusions}

Despite the simplifications in the \flatland environment, the essence of the real-world challenges from railway networks are still well represented in the problem formulation. Algorithmic solutions found for the \flatland benchmark can be integrated into current prototypes for traffic management systems without major modifications. For instance, some of the top methods of the 2019 \flatland benchmark were successfully adapted to real-world internal scheduling and data models by the Swiss Railway (SBB). \flatland instances can be transformed into the SBB-internal data format for further investigation. At Deutsche Bahn (DB) first prototypes pf a planning system were developed on a previous version of \flatland and are now extended to work in a more realistic railway simulation that models the key characteristics of a digitalized railway system of the future.

With \flatland, we introduced a framework that allows to experiment complex VRSP scenarios through OR and RL approaches on a grid world. Our results demonstrate that RL has the potential to effectively and efficiently solve VRSP problems. This is further underlined by the preliminary results from the NeurIPS 2020 \flatland Benchmark that successfully used the approaches from our experiments as well as custom built solutions with even better performance. However, more research is needed to make reinforcement learning a solid alternative to operations research and eventually address problems for which the latter is ill-equipped.

\section{Acknowledgments}

The CCPPO baseline was developed in a joint project of Deutsche Bahn and InstaDeep with a substantial contributions from Alain-Sam Cohen (as.cohen@instadeep.com) from InstaDeep. We would like to thank the efforts from Shivam Khandelwal (AIcrowd) and Jyotish Poonganam (AIcrowd) in helping ensure a stable execution of the competition. We would like to thank Mattias Ljungström for contributions to the many discussions in the initial formative phases of this project. We would like to thank Francois Ramond and the team at SNCF for their continued support for executing this project. We would like to thank our collaborators at NVIDIA for offering prizes for the NeurIPS 2020 Flatland Benchmark. We would like to thank our collaborators at SBB, DB and SNCF for supporting this project with resources and foundational ideals. We would like to thank the AIcrowd community for their continued engagement with this problem.

\printbibliography

\end{document}